\documentclass[letterpaper,twocolumn,10pt]{article}
\usepackage{zhanggroup}

\usepackage{tabularx}
\usepackage{multirow}
\usepackage{tcolorbox}
\tcbset{colback=gray!10, colframe=black, boxrule=0.5pt, arc=2pt}
\usepackage{amsmath}
\usepackage{float}
\usepackage{xspace}

\newcommand{\mypara}[1]{\noindent\textbf{#1}.}

\newcommand{\eg}{e.g.\xspace}

\newcommand{\etal}{et al.\xspace}
\newcommand{\OurMethod}{AIFo\xspace}

\begin{document}

\date{}

\title{\bf From Evidence to Verdict: An Agent-Based Forensic Framework for AI-Generated Image Detection}

\author{
Mengfei Liang\ \ \
Yiting Qu\ \ \
Yukun Jiang\ \ \
Michael Backes\ \ \
Yang Zhang\textsuperscript{}\thanks{Yang Zhang is the corresponding author.}\ \ \
\\
\\
\textit{CISPA Helmholtz Center for Information Security} \ \ \ 
}

\maketitle

\begin{abstract}
The rapid evolution of AI-generated images poses growing challenges to information integrity and media authenticity.
Existing detection approaches face limitations in robustness, interpretability, and generalization across diverse generative models, particularly when relying on a single source of visual evidence.
We introduce \OurMethod (Agent-based Image Forensics), a training-free framework that formulates AI-generated image detection as a multi-stage forensic analysis process through multi-agent collaboration.
The framework integrates a set of forensic tools, including reverse image search, metadata extraction, pre-trained classifiers, and vision-language model analysis, and resolves insufficient or conflicting evidence through a structured multi-agent debate mechanism.
An optional memory-augmented module further enables the framework to incorporate information from historical cases.
We evaluate \OurMethod on a benchmark of 6,000 images spanning controlled laboratory settings and challenging real-world scenarios, where it achieves 97.05\% accuracy and consistently outperforms traditional classifiers and strong vision-language model baselines.
These findings demonstrate the effectiveness of agent-based procedural reasoning for AI-generated image detection.
\end{abstract}

\section{Introduction}
\label{section:intro}

\begin{figure*}[!t]
\centering
\includegraphics[width={1.4\columnwidth}]{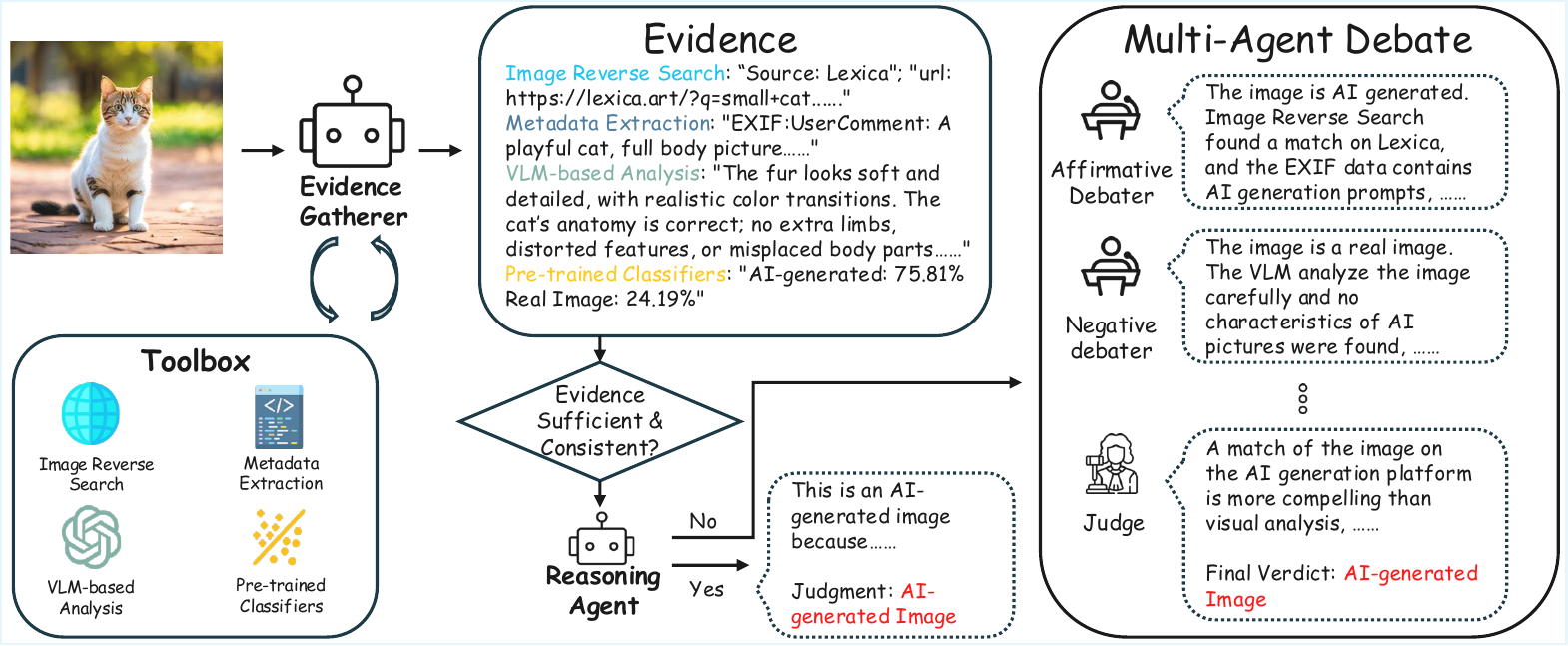}
\caption{High-level overview of our proposed \OurMethod.
}
\label{figure:overview}
\end{figure*}

Recent years have witnessed rapid advancements in image generative models, ranging from early works such as GLIDE~\cite{NDRSMMSC21}, Imagen~\cite{SCSLWDGAMLSHFN22}, DALL·E 2~\cite{RDNCC22}, and Stable Diffusion~\cite{RBLEO22} to more recent foundation models including HunyuanImage~\cite{CCCCCDDGGGo25} and Qwen-Image~\cite{WLZLGYYBXCCTZWYYCLLZMWNCCPQWWYWFXWZZWCL25}.
They can synthesize photorealistic images from natural language in seconds.
However, the realism of AI-generated images has raised serious societal concerns, as they can be used for disinformation, impersonation, and privacy infringement, undermining public trust~\cite{SWAZ23,V20}.
In response to these risks, substantial research has been devoted to the detection of AI-generated images.
Current methodologies can be generally classified into two main categories: traditional machine learning classifiers and approaches leveraging large vision-language models (VLMs).

Traditional machine learning classifiers typically rely on training convolutional neural networks (CNNs) or transformer-based models to distinguish between real and fake images~\cite{WWZOE20,ZXLQZ24,WBZWHCL23,CPNV24,BLYXLHC23}.
Early studies reveal that AI-generated images tend to exhibit shared low-level artifacts, allowing detectors trained on labeled images to identify them~\cite{WWZOE20,SLYZ22}.
For example, DE-FAKE~\cite{SLYZ22} trains a set of classifiers on AI-generated and real images to learn AI-specific artifacts.
While effective under controlled settings, such approaches often struggle to generalize to unseen generative models and provide limited insight into how individual cues contribute to the final decision~\cite{ZLWSJYDSWJ25,WDWJYZZQXZM25,LLWLWRL24}.

More recently, VLMs have shown promise for more generalizable and explainable detection~\cite{ZLWSJYDSWJ25,YFGFXC25,LLWLWRL24,JHZCLZWZZ25}.
Due to the large-scale pre-training, VLMs can be transferred to image detection tasks in a zero-shot or few-shot manner~\cite{ZLWSJYDSWJ25,YFGFXC25,WDWJYZZQXZM25}, without relying on specialized labeled datasets.
Moreover, through prompt engineering, VLMs are able to produce interpretable justifications alongside their predictions~\cite{YFGFXC25,JHZCLZWZZ25,WDWJYZZQXZM25}.

Despite these advancements, both traditional and VLM-based methods remain limited compared to human forensic experts.
Human experts rarely rely on a single visual cue, instead integrating diverse evidence sources and iteratively refining their conclusions as new information emerges.
By contrast, most existing methods perform fixed classification and lack explicit mechanisms to assess evidence sufficiency, resolve conflicting signals, or accumulate experience over time.
Motivated by this, we seek a new paradigm that integrates the strengths of classifiers and VLMs with the procedural reasoning capabilities exhibited by human experts for AI-generated image detection.

Instead of developing another image classifier, we argue that AI-generated image detection should be viewed as an evidence-driven decision-making problem.
Therefore, we adopt a procedural reasoning formulation inspired by human forensic workflows, in which a verdict is reached through multiple stages of reasoning rather than a single forward pass.
We present \OurMethod (Agent-based Image Forensics), a training-free, multi-agent framework that coordinates multiple forensic tools through specialized agent roles.
As shown in \autoref{figure:overview}, \OurMethod structures the detection process into distinct stages, including evidence collection, evidence assessment, and an optional debate stage.
An Evidence Gatherer Agent invokes tools from a forensic toolbox and aggregates their outputs, while a Reasoning Agent evaluates whether the collected evidence is sufficient and consistent to support a reliable decision.
When evidence is ambiguous or contradictory, a debate mechanism is activated, allowing opposing hypotheses to be examined under the supervision of a Judge Agent.
We further introduce a memory module that stores historical cases, enabling the framework to accumulate experience and improve performance over time.
We evaluate \OurMethod on a dataset of 6,000 AI-generated and real images, comprising 3,000 samples from five established benchmarks (\eg, Flickr30K~\cite{PWCCHL15}, GenImage~\cite{ZCYHLLTHHW23}, FakeBench~\cite{LLWLWRL24}, among others) and 3,000 real-world images from six online platforms.
Across these settings, \OurMethod consistently outperforms traditional classifiers such as CNNSpot~\cite{WWZOE20} and DE-FAKE~\cite{SLYZ22}, as well as strong VLM baselines (\eg, GPT series~\cite{O23}), achieving higher accuracy and robustness under common perturbations.

The main contributions of our work are:
(1) We reformulate AI-generated image detection as an evidence-driven forensic decision problem, highlighting the role of evidence sufficiency, consistency, and conflict resolution under uncertainty.
Building on this, we present an agent-based procedural reasoning framework that implements key stages of human forensic analysis, including evidence aggregation, evidence assessment, and structured conflict resolution.
(2) Our proposed framework \OurMethod is a training-free system that integrates conventional classifiers and VLMs as complementary evidence sources rather than standalone detectors, with a modular design that enables robust generalization to evolving generative models.
(3) We conduct extensive empirical evaluations on a benchmark of 6,000 images combining public datasets and real-world samples, and demonstrate the superior performance, generalizability, and robustness of \OurMethod over state-of-the-art baselines.

\section{Related Work}

\mypara{Fake Image Detection}
Early approaches to AI-generated image detection mainly rely on training image classifiers using machine learning, often using datasets produced by specific generative models. 
Over time, research has shifted towards more generalizable and robust detection strategies, including leveraging the visual capabilities of large multimodal models (LMMs) such as CLIP~\cite{RKHRGASAMCKS21}.
CNNSpot~\cite{WWZOE20} proposed one of the first universal detectors independent of generator architecture or dataset. 
De-Fake~\cite{SLYZ22} combined image content and textual prompts using CLIP-based hybrid training.  
DIRE~\cite{WBZWHCL23} and ZeroFake~\cite{STLBZ24} exploit intrinsic differences between real and fake images revealed during the diffusion model reconstruction process to build detection models with improved generalization. 
PatchCraft~\cite{ZXLQZ24} focuses on local texture patches, identifying subtle artifacts left by generative models in fine-grained regions.
AIDE~\cite{YLCHJHX24} detects AI-generated images by selecting highest and lowest frequency patches through a mixture-of-experts architecture.

Recent research leverages VLMs for detection, such as AntifakePrompt~\cite{CYCY23}, Jia \etal~\cite{JLZCYJHLWL24}, and Ji \etal~\cite{JHZCLZWZZ25}. 
AIGI-Holmes~\cite{ZLWSJYDSWJ25} proposes a complete framework to train a VLM for explainable and generalizable detection, aiming to produce human-verifiable justifications.
The work by Yu \etal~\cite{YFGFXC25} develops a framework to enhance generalization and explainability by using a knowledge-guided detector and a forgery-aware prompt learner.
FakeBench~\cite{LLWLWRL24} and DFBench~\cite{WDWJYZZQXZM25} introduced a large-scale benchmark to rigorously test the detection performance of LMMs against a wide range of modern generative models.

Despite these advances, existing methods still mainly rely on internal, pixel-level visual features, overlooking complementary external evidence critical to forensic reasoning. 
In contrast, our agent-based framework emulates human investigative workflows, leverages external tools and reason over different evidence, which allows for a robust and explainable framework for AI-generated image detection.

\mypara{LLM-Based Multi-Agent Frameworks}
Recent advances in LLMs have catalyzed the development of agentic frameworks that simulate complex human-like workflows by coordinating multiple specialized agents.
ReAct~\cite{YZYDSNC23} introduces a framework that switches between reasoning and acting within LLMs.
Other works such as MetaGPT~\cite{HZCZCWZWYLZRXWS24} simulate various roles in a software company and build a multi-agent software development framework.
Moreover, multi-agent frameworks have also been applied to adversarial defense~\cite{ZWZWW24}, harmful content detection~\cite{LFLWD25}, bias identification in generative models~\cite{WBYXCMZCW23}, fake news verification~\cite{LZM24}, and misinformation evaluation~\cite{HYYX25}. 
However, existing approaches primarily emphasize predictive accuracy and explainability, while leaving evidence aggregation and conflict resolution largely implicit.
We therefore present the first training-free multi-agent framework tailored to this task, combining the strengths of existing detectors and LLM agents for adaptive, explainable AI image forensics.

\section{The \OurMethod Framework}
\label{section:framework}

\subsection{Design Rationale}

Instead of treating AI-generated image detection as a single-shot visual classification problem, we formulate it as an evidence-driven reasoning process that operates under heterogeneous, incomplete, and sometimes conflicting evidence, emulating the procedural reasoning of human forensic experts~\cite{FH24, NLBFPCL24}.
Our framework follows a multi-stage reasoning procedure: (1) collect evidence from multiple sources, (2) assess whether the evidence is sufficient or consistent to justify a decision, (3) resolve conflicts through debate mechanism, and (4) produce a verdict with traceable rationale.

\subsection{Overview}
\label{section:evidence_representation}

Given an input image $I \in \mathcal{V}$, the objective is to produce a binary verdict $D \in \mathcal{D}=\{\texttt{AI}, \texttt{Real}\}$ together with a rationale $R \in \mathcal{R}$.
Rather than predicting $D$ directly from pixel-level features, the framework reasons over a set of heterogeneous evidence items $E = \{e_1, e_2, \dots, e_m\}$, where each $e_k$ is an evidence item produced from a distinct evidence channel (e.g., provenance, metadata, model predictions, semantic analysis).
Each evidence item is represented as a structured record $e_k = \langle t_k,\; s_k,\; \xi_k \rangle$, where $t_k$ denotes the evidence type, $s_k \in \{\texttt{AI}, \texttt{Real}, \texttt{Unknown}\}$ indicates the qualitative indication suggested by the evidence source, and $\xi_k$ contains the structured output produced by the corresponding tool. In \OurMethod, the process of collecting evidence is carried out by an Evidence Gatherer Agent, which invokes a set of forensic tools and normalizes their outputs into the unified evidence representation above.

The complete evidence set $E$ then serves as the input to a reasoning module, whose role is to qualitatively assess whether the evidence provides adequate support for a reliable decision and to identify agreements or contradictions across evidence items.
The assessment such as evidence sufficiency and consistency are not computed as predefined metrics, but are instead explicitly reasoned about by a Reasoning Agent based on the available evidence and its contextual descriptions.
Evidence sufficiency refers to whether the currently available evidence provides adequate support for a confident decision, and 
evidence consistency captures the degree to which different evidence sources support compatible conclusions.
If the reasoning module concludes that the evidence is insufficient or contains unresolved conflicts, the framework defers commitment and invokes a structured conflict resolution procedure.
Inspired by previous work~\cite{DLTTM24,LHJWWWYST24,LFLWD25,LDZHGLI24,SGDBP24}, the framework adopts a multi-agent debate (MAD) mechanism to explicitly examine competing arguments rather than aggregating evidence.
The debate is organized around two opposing hypotheses that the input image is AI-generated or real, and two agents are instantiated with complementary roles, each tasked with constructing the argument in support of one hypothesis using the same set of collected evidence.
They exchange arguments over up to \(n\) rounds, refining their reasoning based on prior exchanges at each round, and the debate continues until the supervising Judge Agent determines that one hypothesis is more defensible or the maximum number of rounds is reached.
At termination, the Judge Agent synthesizes the debate history \(H\) together with the original evidence \(E\) to produce a final judgment \(D \in \mathcal{D}\) and a corresponding explanation \(R \in \mathcal{R}\): \(A_J: (H, E) \rightarrow \mathcal{D} \times \mathcal{R}\).
We instantiate the above procedural reasoning framework in \OurMethod, a training-free, LLM-based multi-agent system that maps each stage of the decision process to a specialized agent role.
Each agent operates under a clearly defined role with restricted responsibilities and are guided by structured prompts.

\subsection{Evidence Providers and Tool Interfaces}
\label{section:tools}

\OurMethod uses a modular toolbox of evidence providers. 
The forensic toolbox is a collection of specialized modules that the Evidence Gatherer Agent can invoke to analyze the input image. 
The tools can be broadly categorized into the following four classes.

\mypara{Reverse image search}
Reverse image search tools provide provenance and online distribution of input images by searching for exact or visually similar matches across the internet.
We employ two complementary search tools: one based on the Google Cloud Vision API for high-precision matches, and one web-automation method inspired by Xu et al.~\cite{XML24} to capture a broader set of visually similar results.
Such provenance information can be highly indicative of authenticity, \eg, AI-generated images often appear on generative art platforms, whereas real images are more likely found on news or photography websites. 

\mypara{Metadata Extraction}
This tool extracts and analyzes embedded EXIF metadata (\eg, camera parameters, timestamps, GPS) to identify authenticity markers that differentiate real photographs from AI-generated images. 
We employ ExifTool for selective EXIF metadata extraction, focusing on key camera and capture-related fields while filtering irrelevant noise.
The full list of key fields is provided in \autoref{appendix:metadata_fields}.

\mypara{Pre-trained classifiers}
We integrate multiple publicly available transformer-based detectors as an ensemble (details in \autoref{appendix:hf_models}).
Each model independently predicts the probability of AI generation and outputs an AI-generation score $s_i$.
A final prediction score is obtained via a weighted voting scheme with equal model weights:
\begin{equation}
\text{Prediction Score} = \frac{\sum_{i=1}^{N} w_i \cdot s_i}{\sum_{i=1}^{N} w_i},
\end{equation}
where $w_i$ represents the weight of the $i$-th model $\theta_i$, $s_i$ is the AI confidence score from $\theta_i$, and $N$ is the total number of loaded models. 
We use equal weights for simplicity and this ensemble design enhances robustness and reduces overfitting to specific generators during inference.

\mypara{VLM-based analysis}
A VLM-based tool conducts semantic image analysis to detect visual and contextual cues indicative of AI synthesis, examining artifacts such as lighting inconsistencies, unnatural textures, anatomical anomalies, and implausible contextual relationships.
For real images, it identifies coherent structural, lighting, and contextual patterns. 
Beyond binary classification, the tool produces interpretable textual explanations and confidence estimates, enhancing the transparency and reliability of the judgments.

\section{Evaluation}

\subsection{Dataset Construction}
\label{section:dataset}

\begin{table*}[!t]
\centering
\caption{Performance comparison of different methods on our benchmark dataset comprising three evaluation subsets: \textit{Overall}, \textit{In-the-Lab}, and \textit{In-the-Wild}. 
Metrics reported are Accuracy (Acc), Precision (Prec), Recall (Rec), and F1-score (F1). 
Best results are highlighted in \textbf{bold} and second best results are \underline{underlined}.}
\label{table:metrics}
\setlength{\tabcolsep}{3pt}
\scalebox{0.85}{
\begin{tabular}{lcccc | lcccc | lcccc}
\toprule
\multicolumn{5}{c|}{\textbf{Overall}} & \multicolumn{5}{c|}{\textbf{In-the-Lab}} & \multicolumn{5}{c}{\textbf{In-the-Wild}} \\
\cmidrule(lr){1-5} \cmidrule(lr){6-10} \cmidrule(lr){11-15}
\textbf{Method} & \textbf{Acc} & \textbf{Prec} & \textbf{Rec} & \textbf{F1} &
 & \textbf{Acc} & \textbf{Prec} & \textbf{Rec} & \textbf{F1} &
 & \textbf{Acc} & \textbf{Prec} & \textbf{Rec} & \textbf{F1} \\
\midrule
CNNSpot~\cite{WWZOE20}  & 0.5277 & 0.9826 & 0.0563 & 0.1066 &
         & 0.5553 & 0.9882 & 0.1120 & 0.2012 &
         & 0.5000 & 0.5000 & 0.0007 & 0.0013 \\
PatchCraft~\cite{ZXLQZ24} & 0.6517 & 0.7423 & 0.4647 & 0.5715 &
         & 0.8123 & 0.8704 & 0.7340 & 0.7964 &
         & 0.4910 & 0.4780 & 0.1953 & 0.2773 \\
DE-FAKE~\cite{SLYZ22}  & 0.7142 & 0.6820 & 0.8027 & 0.7374 &
         & 0.6720 & 0.6673 & 0.6860 & 0.6765 &
         & 0.7563 & 0.6933 & \underline{0.9193} & 0.7905 \\
GPT-4o~\cite{O23}  & \underline{0.9483} & \underline{0.9920} & \underline{0.9038} & \underline{0.9458} &
         & \underline{0.9537} & \textbf{0.9938} & \underline{0.9130} & \underline{0.9517} &
         & \underline{0.9428} & \underline{0.9900} & 0.8947 & \underline{0.9399} \\
AIFo (ours)  & \textbf{0.9705} & \textbf{0.9920} & \textbf{0.9487} & \textbf{0.9698} &
         & \textbf{0.9740} & \underline{0.9917} & \textbf{0.9560} & \textbf{0.9735} &
         & \textbf{0.9670} & \textbf{0.9923} & \textbf{0.9413} & \textbf{0.9661} \\
\bottomrule
\end{tabular}
}
\end{table*}

To comprehensively evaluate our multi-agent framework under both controlled and open-world conditions, we construct a benchmark dataset covering two complementary settings: \textit{in-the-lab} and \textit{in-the-wild}.
The \textit{in-the-lab} setting consists of well-curated datasets commonly used in prior work, where image sources and generation processes are relatively controlled.
In contrast, the \textit{in-the-wild} setting comprises images collected from unconstrained online platforms, reflecting the diversity, noise, and uncertainty encountered in real-world forensic scenarios.

The benchmark contains 6,000 images in total, evenly split between the two settings, with 1,500 real and 1,500 AI-generated images per setting.
For the \textit{in-the-lab} setting, real images are sampled from Flickr30k~\cite{PWCCHL15}, ImageNet~\cite{DDSLLF09}, and DIV2K~\cite{AT17} (500 images each).
AI-generated images are obtained from GenImage~\cite{ZCYHLLTHHW23} and FakeBench~\cite{LLWLWRL24}, by sampling 100 images from each of the eight models in GenImage and 70 images from each of the ten models in FakeBench, yielding 1,500 AI-generated samples.

For the \textit{in-the-wild} subset, real images are collected from publicly available online sources, including Flickr, Wikimedia Commons, and the Holopix50k dataset~\cite{HKURGOL20}.
To ensure diversity, we select ten keywords: \textit{animal}, \textit{building}, \textit{food}, \textit{indoor}, \textit{landscape}, \textit{person}, \textit{plant}, \textit{snow}, \textit{sport}, \textit{transportation} and \textit{water}. 
Images from Flickr and Wikimedia Commons are retrieved via keyword search and randomly sampled from the results.
For Holopix50k, which lacks explicit semantic labels, we use BLIP~\cite{LLXH22} to categorize images into the same keyword set before random sampling.
For each keyword and each source, we sample 50 images, yielding a balanced collection of real-world photographs.
The corresponding AI-generated images are sourced from three popular generative art platforms: Lexica, NightCafe, and Civitai.
Images from these platforms are generated by a wide range of text-to-image models, including DALL·E~\cite{RDNCC22}, Stable Diffusion~\cite{RBLEO22}, SDXL~\cite{PELBDMPR23}, StyleGAN~\cite{KLAHLA20}, and numerous community fine-tuned variants.
We use the same set of ten keywords to retrieve AI-generated images and randomly sample 50 images per keyword from each platform.
As summarized in \autoref{appendix:ai_model_sources}, the resulting benchmark spans over 20 generative models, enabling evaluation under both controlled and open-world conditions.

\subsection{Experimental Setup}

\mypara{Implementation Details.}
We evaluate \OurMethod under a unified decision protocol that reflects its full procedural reasoning pipeline.
The framework is implemented using LangGraph, which supports stateful multi-agent workflows.
All agents share the same LLM backbone (\eg, GPT-4o), ensuring consistent reasoning capacity across different agent roles. 
Unless otherwise specified, all agents are instantiated with temperature set to 0 and a fixed random seed (42) to reduce stochastic variation in reasoning traces.
All external tool outputs, including reverse image search results and metadata extraction, were cached at evaluation time and treated as fixed inputs during benchmarking to ensure reproducibility.

\mypara{Evaluation Metrics.}
We evaluate performance using standard binary classification metrics, including \textbf{Accuracy}, \textbf{Precision}, \textbf{Recall}, and \textbf{F1-score}, where AI-generated images are treated as positive samples.
While these metrics provide a high-level summary of detection effectiveness, they do not capture how or why a decision is reached.
We therefore include targeted analyses that examine intermediate decision dynamics, such as when additional reasoning is triggered and how often the final verdict differs from the initial assessment.

\mypara{Baseline Methods.}
We compare \OurMethod against representative baseline approaches spanning both conventional classifiers and VLMs.
For classifier-based methods, we include CNNSpot~\cite{WWZOE20}, DE-FAKE~\cite{SLYZ22}, and PatchCraft~\cite{ZXLQZ24}.
For VLM baselines, we adopt the same LLM backbone as used by the agents (\eg, GPT-4o) to ensure a fair comparison.
Each VLM receives the input image together with a binary classification prompt.
We experiment with multiple prompt techniques for the VLM baselines such as Chain-of-Thought (CoT)~\cite{WWSBIXCLZ22} and report the best results, to provide a competitive reference for evaluating the effectiveness of our framework.

\subsection{Results}

\mypara{Overall Performance}
\autoref{table:metrics} reports the overall performance of \OurMethod and baseline approaches across three evaluation settings.
The results in this table are obtained using GPT-4o as the shared LLM backbone for all agents.
These metrics summarize the outcome of the complete decision process, in which a verdict may be produced either directly from initial evidence assessment or after invoking conflict-aware procedural reasoning.
A detailed per-setting breakdown is provided in \autoref{appendix:detailed_accuracy}.
Traditional classifiers (CNNSpot~\cite{WWZOE20}, DE-FAKE~\cite{SLYZ22}, PatchCraft~\cite{ZXLQZ24}) perform poorly, with limited ability to generalize beyond curated datasets.
GPT-4o achieves strong overall accuracy but tends to rely predominantly on visual semantics, leading to lower recall in cases where AI-generated images closely mimic real photographic patterns.
Across all evaluation settings, \OurMethod achieves competitive or superior overall performance compared to all baselines.
In the full evaluation, it attains an accuracy of 0.9705 and an F1-score of 0.9698, exceeding GPT-4o by 2.22\% in accuracy and 2.40\% in F1-score.

\begin{figure*}[!t]
\centering
\includegraphics[width={0.85\textwidth}]{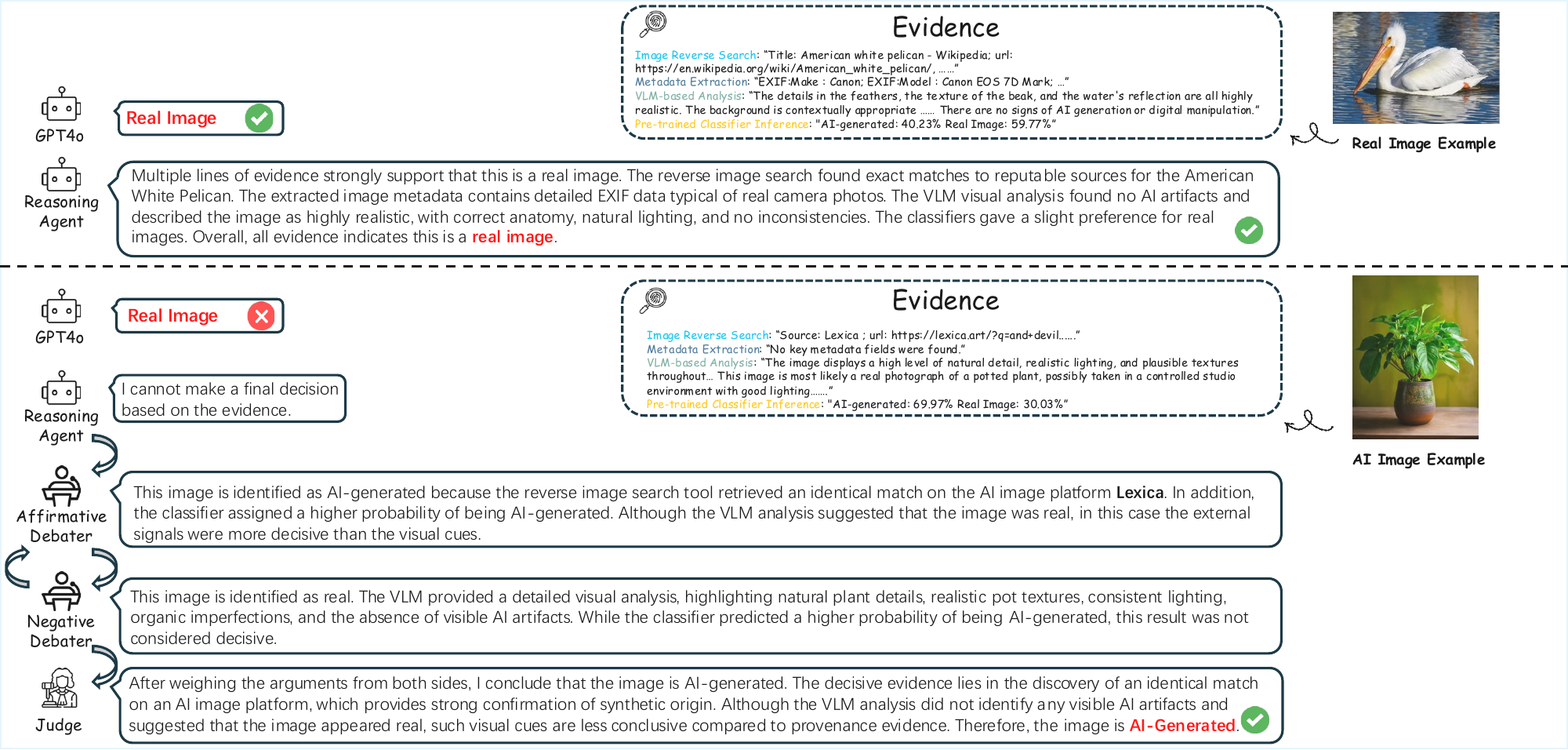}
\caption{Examples of our agent framework's decision-making process.}
\label{figure:example}
\end{figure*}

\begin{table}[!t]
\centering
\caption{Ablation Study of AIFo Components with GPT-4o Backbone}
\label{table:ablation}
\setlength{\tabcolsep}{4pt}
\small
\begin{tabular}{lcccc}
\toprule
\textbf{Method} & \textbf{Acc} & \textbf{Prec} & \textbf{Rec} & \textbf{F1} \\
\midrule
GPT-4o (direct) & 0.9483 & 0.9920 & 0.9038 & 0.9458 \\
GPT-4o + CoT & 0.9510 & 0.9906 & 0.9107 & 0.9489 \\
\midrule
AIFo w/o Tools & 0.9525 & 0.9913 & 0.9130 & 0.9505 \\
AIFo w/o MAD & 0.9635 & \textbf{0.9922} & 0.9343 & 0.9624 \\
AIFo & \textbf{0.9705} & 0.9920 & \textbf{0.9487} & \textbf{0.9698} \\
\bottomrule
\end{tabular}
\end{table}

\mypara{Quantitative and Qualitative Analysis}
To understand where the performance gains of \OurMethod originate, we analyze test samples misclassified by GPT-4o but correctly classified by \OurMethod. 
Among 6,000 samples, \OurMethod correctly identifies over 140 additional AI-generated images. 
Approximately 64\% of these corrections are attributed to external evidence unavailable to single VLM inference. 
Roughly 34\% of the corrected samples involve insufficient or conflicting signals across different evidence channels, which trigger the multi-agent debate process.
In such cases, initial assessments are revised through structured evaluation of competing hypotheses, yielding a corrected final verdict by the Judge Agent.
\autoref{table:ablation} further shows that removing either external evidence tools or the multi-agent debate mechanism consistently degrades performance compared to the full framework, confirming that both components contribute meaningfully to the overall gains.

Qualitative examples in \autoref{figure:example} further illustrate this process. 
In the first real-image case, multiple evidence sources provide consistent support for authenticity, allowing the framework to reach a direct verdict without invoking further conflict resolution stage.
In the second case, the realistic appearance misleads the baseline GPT-4o and the VLM tool.
However, conflicting provenance and classifier evidence induce the framework to defer commitment and invoke multi-agent debate.
Through structured argumentation, the system identifies the inconsistency in the assessment and ultimately produces the correct AI-generated verdict.

Overall, this analysis shows that the gains of \OurMethod come from two complementary factors: the ability to override misleading visual cues using stronger non-visual evidence, and the capacity to revise initial judgments through explicit reasoning when evidence is insufficient or contradictory.

\begin{figure}[!t]
\centering
\includegraphics[width={0.9\columnwidth}]{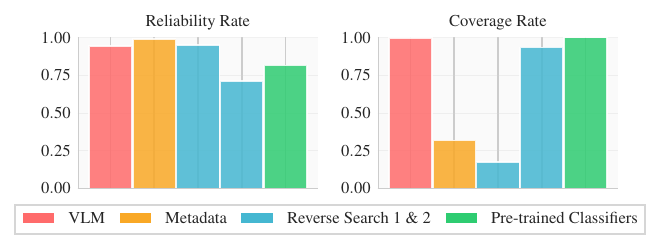}
\caption{Analysis of individual tool contributions to the agent framework.}
\label{figure:tool_analysis}
\end{figure}

\begin{figure}[!t]
\centering
\includegraphics[width={1\columnwidth}]{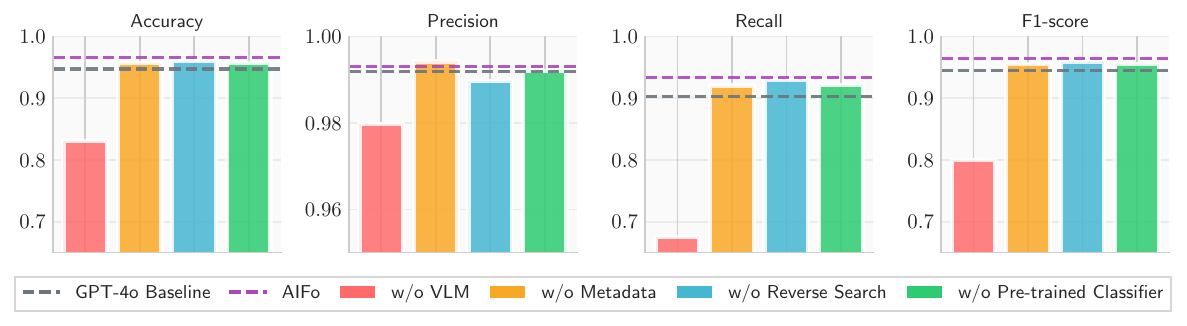}
\caption{Performance degradation when each tool is disabled from the framework.}
\label{figure:leave-one-out}
\end{figure}

\mypara{Tool Reliability and Decision Pattern Analysis}
To better understand the internal decision-making processes of our framework, we analyze the reliability and coverage of individual forensic tools. 
We define \textit{reliability} as the proportion of cases where a tool’s evidence aligns with the agent’s final decision, and \textit{coverage} as the proportion of decisions for which a tool provides valid evidence.
As shown in \autoref{figure:tool_analysis}, metadata extraction and the first reverse search tool achieve the highest reliability, indicating that their evidence strongly influences final decisions when available.
However, their relatively low coverage indicates that such decisive evidence is only present in a subset of cases.
These tools therefore function as high precision but sparse evidence sources.
In contrast, the VLM-based analysis, pre-trained classifiers, and the second reverse search tool provide near-complete coverage across the dataset.
Among them, VLM-based analysis contributes the most essential visual evidence and serves as the primary source of information in the majority of cases.
This observation is further supported by a leave-one-out analysis shown in \autoref{figure:leave-one-out}, where removing the VLM tool leads to the most pronounced performance degradation, with accuracy dropping below 0.85 and recall below 0.70.
By comparison, disabling metadata extraction, reverse image search, or pre-trained classifiers results in only modest performance degradation, indicating that these tools provide complementary rather than dominant signals.
Instead of relying on any single tool, \OurMethod leverages procedural reasoning to balance these heterogeneous signals.
Highly reliable but low coverage evidence is emphasized when present, while high coverage but noisier evidence provides contextual support when stronger signals are unavailable.
This behavior emerges from the reasoning process itself and is not explicitly hard-coded into the system.

\mypara{Inference Efficiency and Cost Analysis}
The procedural reasoning framework incurs higher computational cost than VLM baselines due to multi-stage reasoning, multi-agent interaction, and external tool invocation.
As summarized in \autoref{table:latency_token}, \OurMethod requires an average of 40.08\,s and 5.2k tokens per image, corresponding to approximately $7.5\times$ and $7.3\times$ the cost of GPT-4o, respectively.
While this overhead is non-negligible, it should be interpreted in the context of the framework’s adaptive decision process.
As shown in earlier analyses, only a subset of samples require conflict resolution reasoning, while the majority are resolved after the initial evidence assessment.
More importantly, the additional computation enables capabilities that single-shot visual reasoning cannot provide.
By integrating provenance, metadata, and model-based evidence, \OurMethod offers verifiable decision rationales and improved reliability in ambiguous cases, which are critical in high-stakes applications.
These results highlight a fundamental trade-off between efficiency and decision robustness, and suggest that our procedural reasoning is most appropriate when interpretability and reliability are prioritized over real-time constraints.

\begin{table}[!t]
\centering
\caption{Average inference latency and token usage per image.}
\label{table:latency_token}
\setlength{\tabcolsep}{1.5pt}
\scalebox{0.85}{
\begin{tabular}{lcc}
\toprule
\textbf{Method} & \textbf{Avg. Latency} & \textbf{Avg. Token Usage} \\
\midrule
GPT-4o & 5.31s & 715.05 \\
AIFo (w/o MAD) & 25.43s & 2728.29 \\
AIFo & 40.08s & 5230.86 \\
\bottomrule
\end{tabular}
}
\end{table}

\begin{table}[!t]
\centering
\caption{\OurMethod vs. GPT-4o performance under image perturbations. \OurMethod’s superior results are in \textbf{bold}.}
\label{table:robustness}
\setlength{\tabcolsep}{1.5pt}
\scalebox{0.85}{
\begin{tabular}{lcccc|cccc}
\toprule
\multirow{2}{*}{} & \multicolumn{4}{c|}{\textbf{GPT-4o}} & \multicolumn{4}{c}{\textbf{AIFo (Ours)}} \\
\cmidrule(lr){2-5} \cmidrule(lr){6-9}
& \textbf{Acc} & \textbf{Prec} & \textbf{Rec} & \textbf{F1} & \textbf{Acc} & \textbf{Prec} & \textbf{Rec} & \textbf{F1} \\
\midrule
Blu & 0.8818 & 0.9662 & 0.7913 & 0.8701 & \textbf{0.9047} & 0.9380 & \textbf{0.8667} & \textbf{0.9009} \\
Noi & 0.9462 & 0.9866 & 0.9047 & 0.9438 & \textbf{0.9690} & \textbf{0.9879} & \textbf{0.9497} & \textbf{0.9684} \\
Sha & 0.9410 & 0.9926 & 0.8887 & 0.9377 & \textbf{0.9670} & 0.9902 & \textbf{0.9433} & \textbf{0.9662} \\
\bottomrule
\end{tabular}
}
\end{table}

\subsection{Robustness Analysis}

To evaluate robustness of our framework under realistic degradations, we test its performance on perturbed versions of the dataset. 
Three common distortions are applied: Gaussian blur (radius = 2) to simulate defocus, Gaussian noise (mean = 0, variance = 2) to mimic sensor noise, and sharpening (factor = 2.0) to simulate edge distributions. 
These perturbations reflect typical degradations in real-world imaging and compression pipelines and we compare the performance against the best baseline, GPT-4o.

As shown in \autoref{table:robustness}, both \OurMethod and GPT-4o experience performance degradation under blur due to the loss of fine-grained visual details.
However, \OurMethod consistently maintains higher accuracy across all perturbation types.
We observe that perturbations increase disagreement between visual evidence and other evidence sources, thereby triggering debate mechanism more frequently.
This allows the framework to rely less on degraded visual cues and instead emphasize complementary non-visual evidence when available.
Under noise and sharpening, where visual distortions are less severe, \OurMethod exhibits only minor performance drops relative to the clean setting.
This behavior suggests that procedural reasoning provides a mechanism for adapting to changes in evidence reliability.
Overall, these results indicate that robustness in \OurMethod emerges from its ability to mediate heterogeneous evidence under varying degrees of visual uncertainty.

To evaluate the resilience of \OurMethod against evasive attacks, we simulate two representative attack scenarios that target the framework's key evidence sources. 
The first attack employs reverse image search manipulation techniques and the second attack involves metadata forgery.
In the first scenario, we manipulate the provenance evidence returned by the reverse image search tool.
Specifically, we use GPT-4o to generate counterfactual search results for each image. 
For real images, we fabricate search results indicating that the image was sourced from an AI generation platform, while for AI-generated images, we create results suggesting the image originated from a reputable photography website.
These results are then injected into the evidence set returned to the agent.
In the second scenario, we perform metadata forgery by swapping EXIF metadata between AI-generated and real images.
Real images are randomly assigned metadata extracted from AI-generated samples, while AI-generated images receive real images' metadata.

As shown in \autoref{table:Evasive_attack}, \OurMethod experiences a moderate performance degradation under both evasive attack settings.
This degradation is primarily due to the framework's limitation in verifying the authenticity of evidence returned by external tools.
Since \OurMethod is designed to treat tool outputs as trustworthy forensic sources, deliberately falsified information can mislead the agent, resulting in false judgments.
To enhance robustness against such attacks, several potential defenses can be considered.
Firstly, implementing cross-tool consistency validation can help identify conflicting evidence that may indicate manipulation such as verifying that metadata timestamps and reverse-search provenance sources are mutually coherent.
Secondly, trust-weighted evidence aggregation can be introduced, where each tool's output is dynamically weighted based on historical reliability.
Finally, incorporating external verification layers such as digital watermark authentication can help validate evidence before feeding it into the reasoning pipeline.
These strategies would allow \OurMethod\ to better distinguish adversarially manipulated evidence, thereby improving its resilience against real-world evasive attacks.

\begin{table}[!t]
\centering
\caption{Performance of \OurMethod under two evasive attack scenarios.}
\label{table:Evasive_attack}
\setlength{\tabcolsep}{1.5pt}
\scalebox{0.85}{
\begin{tabular}{lcccc}
\toprule
\textbf{Attack Type} & \textbf{Acc} & \textbf{Prec} & \textbf{Rec} & \textbf{F1} \\
\midrule
None (clean) & 0.9705 & 0.9920 & 0.9487 & 0.9698 \\
Reverse search manipulation & 0.8971 & 0.8690 & 0.9353 & 0.9010 \\
Metadata forgery & 0.8702 & 0.8399 & 0.9147 & 0.8757 \\
\bottomrule
\end{tabular}
}
\end{table}

\subsection{Takeaway}
\label{section:takeaway}

This evaluation demonstrates that procedural reasoning offers distinct advantages for AI-generated image detection, particularly in scenarios where evidence is incomplete, ambiguous, or conflicting.
By explicitly assessing evidence sufficiency and resolving conflicts through structured reasoning, the framework is able to revise misleading initial judgments and produce more stable and interpretable decisions.
At the same time, these benefits come with increased computational cost, underscoring a fundamental trade-off between efficiency and decision robustness.
Our results suggest that procedural, agent-based reasoning is most appropriate for forensic and high-stakes applications, where reliability, and auditability are prioritized over real-time inference.
Together, these findings support that AI-generated image detection is better viewed as an evidence-driven decision process rather than a purely predictive task, and highlight procedural reasoning as a promising direction for addressing the limitations of existing detectors.

\section{Memory-Augmented Reasoning}
\label{section:memory_reasoning}

\subsection{Motivation and Design}

The core procedural reasoning framework operates in a stateless manner, analyzing each image independently without access to prior cases.
While this design ensures fairness and prevents information leakage in benchmark evaluation, human forensic reasoning is often cumulative, drawing on past experiences to inform judgments in difficult or ambiguous cases~\cite{HRMP25}.
Motivated by this observation, we explore a memory-augmented reasoning module as an optional component that provides contextual reference to past cases.
The memory module maintains a knowledge base of previously analyzed cases, including their visual embeddings, collected evidence, final decisions, and reasoning traces.
For failure cases, it additionally stores structured reflections generated by LLM that describe potential causes of error, such as unreliable evidence sources or misinterpretation of visual cues.
During inference, the memory module retrieves semantically similar cases using CLIP-based embedding similarity.
These retrieved cases are passed to the reasoning agent as supplementary context, allowing it to reflect on similarities and differences between the current input and past cases.
The agent may choose to incorporate this information when reassessing evidence reliability or resolving conflicts, but is not required to follow past decisions.
This design reduces the risk of systematic bias or error propagation, as memory serves as a reference instead of a prescriptive rule.

To avoid data leakage, the knowledge base is constructed from a separate set of 600 images (300 real, 300 AI-generated) distinct from the benchmark dataset.
Each image is embedded using CLIP-ViT-B/32~\cite{RKHRGASAMCKS21}, and indexed for similarity retrieval.
The memory module is integrated as an additional forensic tool within the framework and activated during evidence collection.
When analyzing a new image, it retrieves the top-$k$ most relevant past cases ($k=1$ by default), which are passed to the reasoning agent as additional context.

\subsection{Results Analysis}
\begin{figure}[!t]
\centering
\includegraphics[width={1\columnwidth}]{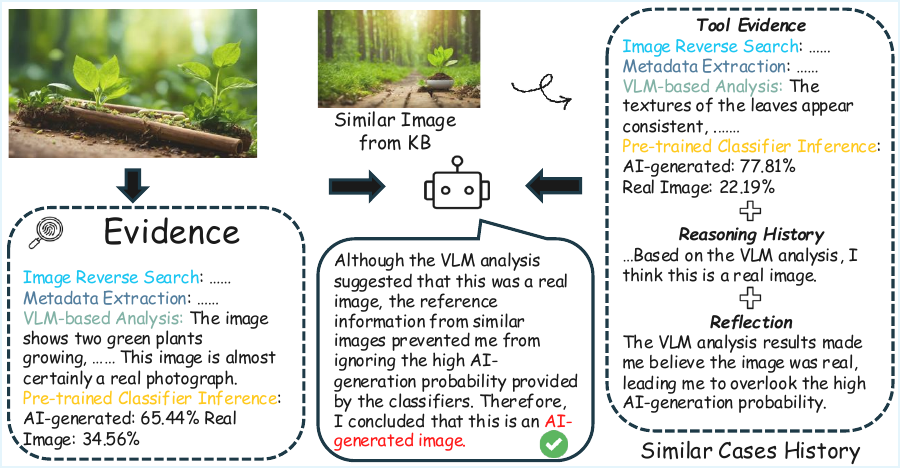}
\caption{Example of memory-augmented reasoning illustrating the impact of similar historical cases on decision-making.}
\label{figure:memory_example}
\end{figure}

We do not include the memory-augmented module in the main evaluation, as constructing the memory knowledge base requires access to ground-truth labels during the experience accumulation phase.
Including such label dependent information at test time would introduce an unfair advantage compared to baseline methods.
Instead, we evaluate the memory module through a controlled failure recovery analysis, which reflects its intended use as an additional mechanism for difficult cases.
Specifically, we analyze 50 misclassified images from the main benchmark for which semantically similar cases exist in the knowledge base.
These cases represent scenarios where the stateless reasoning fails and additional contextual reference may be beneficial.
After enabling memory-augmented reasoning, over 40\% of these previously misclassified samples are corrected (\autoref{table:memory_results}).
Qualitative examples in \autoref{figure:memory_example} illustrate how retrieved past cases support reflective reasoning.
These observations suggest that the memory-augmented reasoning offers substantial benefits in failure recovery and strong potential for enhancing adaptive reasoning.

\begin{table}[!t]
\centering
\caption{Number of errors before and after incorporating similar case history.
FP: False Positive Cases; FN: False Negative Cases.}
\label{table:memory_results}
\setlength{\tabcolsep}{3pt}
\scalebox{0.85}{
\begin{tabular}{lcccccc}
\toprule
& \multicolumn{2}{c}{\textbf{\textit{In-the-Lab}}} & \multicolumn{2}{c}{\textbf{\textit{In-the-Wild}}} & \\
\cmidrule(lr){2-3} \cmidrule(lr){4-5}
& \textbf{FP} & \textbf{FN} & \textbf{FP} & \textbf{FN} & \textbf{Total} \\
\midrule
\textbf{Before} & 2 & 12 & 1 & 35 & \textbf{50} \\
\textbf{After} & 1 & 6 & 0 & 22 & \textbf{29} \\
\bottomrule
\end{tabular}
}
\end{table}

\section{Discussion and Limitations}

Our \OurMethod framework potentially represents a paradigm shift in AI-generated image detection by emulating human forensic reasoning through multi-agent collaboration.
The framework's training-free nature and cross-model generalizability address key limitations of existing detection methods, offering a more sustainable and adaptable solution for the rapidly evolving landscape of generative AI.

Despite the promising results, our approach has several important limitations that warrant careful consideration.

\mypara{Scalability and Computational Efficiency}
While our multi-agent approach achieves high accuracy, it comes with increased computational overhead compared to single-model solutions.
The sequential and parallel execution of multiple forensic tools, combined with LLM-based reasoning, results in higher latency and resource consumption.
For large-scale deployment scenarios, optimization strategies such as tool prioritization, caching mechanisms, and selective tool activation based on image characteristics could help balance accuracy and efficiency.

\mypara{Dependency on External Services}
The framework's effectiveness is inherently tied to the availability and reliability of external services, particularly reverse image search APIs and metadata extraction tools.
Changes in API policies, service outages, or modifications to search algorithms could impact the framework's performance.
This dependency creates potential points of failure that are beyond the system's direct control.

\mypara{Adversarial Metadata Manipulation}
The framework’s reliance on image EXIF metadata as a key source of forensic evidence is a limitation.
However, adversaries could potentially manipulate image metadata to mislead the detection system.
For instance, attackers could inject fake EXIF data mimicking legitimate camera parameters into AI-generated images.
Such metadata spoofing attacks could compromise the reliability of our metadata extraction tool, which currently shows high reliability rates in our evaluation.

\section{Conclusion}

This work presents \OurMethod, a multi-agent procedural reasoning framework that takes AI-generated image detection as an evidence-driven forensic decision process.
Instead of proposing yet another detector, we show that explicitly modeling how forensic evidence is collected, evaluated, and reconciled leads to more reliable and interpretable decisions. 
Extensive experiments demonstrate that integrating heterogeneous evidence within a structured reasoning pipeline enables effective conflict resolution and decision revision.
Our results suggest that robustness in AI-generated image detection is better achieved through multi-stage reasoning over evidence than through increasingly complex detection models alone. 
We believe that procedural, agent-based reasoning provides a promising foundation for developing forensic systems that remain effective as generative models continue to evolve.

\mypara{Limitation}
The proposed framework has higher computational cost than single-model baselines due to multi-stage reasoning and external tool invocation.
While our analysis shows that conflict resolution reasoning is only required for a subset of inputs, further optimization will be important for large-scale deployment.
In addition, parts of the reasoning process, including the assessment of evidence sufficiency and consistency, rely on qualitative judgments produced by LLMs and may introduce variability across models or prompts.
Exploring hybrid approaches that combine procedural reasoning with more formal uncertainty estimation and robustness guarantees is a promising avenue for future research.

\section*{Ethical Considerations}

In this study, we adopted a stakeholder-oriented perspective to examine the ethical dimensions of our work.
For the research team, the development and validation of our new detection framework contributed to advancing technical expertise and academic reputation. 
For the general public, the framework offers a practical tool to mitigate the spread of misinformation by improving the detection of AI-generated images, thereby safeguarding individuals from being misled. 
Companies such as social media platforms and news organizations may also benefit by employing the framework to verify content authenticity and maintain the credibility of their services.  
Our research is guided by several core ethical principles. First, the principle of beneficence is reflected in our aim to protect society from the harmful consequences of misinformation. 
Second, respect for persons is ensured by using only publicly available datasets that do not involve personal or sensitive information. 
Third, the principle of justice informed our effort to design a framework whose outcomes can be applied broadly across different social groups, thereby promoting fair access to reliable information.
We think that this research provides substantial value in promoting information authenticity and strengthening public trust. 
We are therefore confident that the study is ethically sound and makes a meaningful contribution to the ongoing development of AI-generated image detection.

{
\small
\bibliographystyle{plain}
\bibliography{main}
}

\newpage
\appendix

\section{Prompts Used in Our Framework}
\label{appendix:prompts}

This appendix documents the exact prompts used in our experiments to facilitate reproducibility. 
We group prompts according to their roles in the multi-agent framework.

\subsection{Evidence Gatherer Agent Prompt}

The Evidence Gatherer Agent collects cross-source forensic signals. 
\autoref{table:evidence_gatherer_prompt} is the prompt template used to instruct the LLM:

\begin{table}[h!]
\centering
\begin{tcolorbox}[width=0.95\linewidth]
\small

You are an AI Image Forensics Expert. Your task is to determine whether the input image is AI-generated or real using the available forensic tools. \\

- A real image refers to images created by humans, including photographs captured by cameras, photos that have been edited with software such as Photoshop, or human artistic creations such as hand-drawn sketches and paintings. \\
- An AI-generated image refers to images that are fully or partially generated by AI models. \\

Available Tools: \\
- reverse\_search: Perform a reverse image search to find exact matches or similar appearances online. \\
- extract\_image\_metadata: Inspect technical EXIF metadata for authenticity cues. \\
- vlm\_analysis: Obtain expert-level visual analysis of the image content. \\
- pre-trained\_classifiers: Apply dedicated AI-generated image detection models. \\

Your role is to systematically invoke these tools as needed and collect evidence that will later be assessed to determine the authenticity of the input image.
\end{tcolorbox}
\caption{Prompt template for the Evidence Gatherer Agent.}
\label{table:evidence_gatherer_prompt}
\end{table}

\subsection{Reasoning Agent Prompt}

The Reasoning Agent first assesses whether the evidence is sufficient and consistent to support a decision.
If so, it synthesizes all sources to produce a final binary judgment with a explanation that evaluates source reliability.
\autoref{table:reasoning_agent_prompt1} and \autoref{table:reasoning_agent_prompt2} are the prompt templates used to instruct the LLM:

\begin{table}[h!]
\centering
\begin{tcolorbox}[width=0.95\linewidth]
\small
You are an AI Image Forensics Expert. Your task is to determine if the following evidence collected from multiple tools is sufficient and consistent enough to make a final judgment. \\

\{tool\_results\} \\

Answer 'True' if the evidence is both sufficient and consistent enough to confidently reach a final decision and 'False' if the evidence is incomplete, ambiguous, or contains major conflicts that require further debate and analysis. \\
\end{tcolorbox}
\caption{First prompt template for the Reasoning Agent.}
\label{table:reasoning_agent_prompt1}
\end{table}

\begin{table}[h!]
\centering
\begin{tcolorbox}[width=0.95\linewidth]
\small
You are an AI Image Forensics Expert. Your task is to determine whether the image ia AI-generated or a real image. \\

- A real image refers to images created by humans, including photographs captured by cameras, photos that have been edited with software such as Photoshop, or human artistic creations such as hand-drawn sketches and paintings. \\
- An AI-generated image refers to images that are fully or partially generated by AI models. \\

Please make a final judgment based on the following evidence collected from multiple tools: \\

\{tool\_results\} \\

Critically evaluate each evidence source and its reliability. \\

Required output format: \\
1. is\_ai\_generated: boolean (True if AI-generated, False if real image) \\
2. analysis\_details: A detailed analysis explaining your decision \\
\end{tcolorbox}
\caption{Second prompt template for the Reasoning Agent.}
\label{table:reasoning_agent_prompt2}
\end{table}

\subsection{Debate Agents Prompt}
The Debate Agents engage in a structured debate to resolve conflicts and ambiguities in the evidence.
\autoref{table:pro_agent_prompt} is the prompt template used to instruct the Pro-Agent LLM:

\begin{table}[h!]
\centering
\begin{tcolorbox}[width=0.95\linewidth]
\small
You are an AI Image Forensics Expert. Your goal is to correctly classify an image as either AI-generated or real. \\
Your analysis must be based on the evidence provided in the tool results below. \\

Tool Results: \\
\{tool\_results\} \\

\#\textbf{First Round Only:} \\
You are arguing in favor of the image being AI-generated. \\
Scrutinize the tool results for any artifacts, inconsistencies, or patterns typical of AI generation. Present your findings as a concise, bullet-pointed list. Focus on the strongest pieces of evidence that support your assigned perspective. \\

\#\textbf{Subsequent Rounds Only:} \\
Review the other expert's points from the previous round and re-evaluate your own position. \\
- Acknowledge any valid points they made. \\
- Re-examine the tool results to see if their perspective reveals something you missed. \\
- Refine or strengthen your analysis based on this new information. Your updated analysis should be more nuanced. \\

You are arguing in favor of the image being AI-generated. \\
The other expert's (arguing for "Real") points: \\

\{negative\_history\} \\

Provide your updated, refined analysis as a concise bullet-pointed list. \\
\end{tcolorbox}
\caption{Prompt template for the Pro-Agent.}
\label{table:pro_agent_prompt}
\end{table}

\autoref{table:con_agent_prompt} is the prompt template used to instruct the Con-Agent LLM:

\begin{table}[h!]
\centering
\begin{tcolorbox}[width=0.95\linewidth]
\small
You are an AI Image Forensics Expert. Your goal is to correctly classify an image as either AI-generated or real. \\
Your analysis must be based on the evidence provided in the tool results below. \\

Tool Results: \\
\{tool\_results\} \\

\textbf{First Round Only:} \\
You are arguing in favor of the image being authentic (real). \\
Look for signs of naturalness, photographic properties, and details that are hard for AI to replicate, based on the tool results. \\

\textbf{Subsequent Rounds Only:} \\
Review the other expert's points from the previous round and re-evaluate your own position. \\
- Acknowledge any valid points they made. \\
- Re-examine the tool results to see if their perspective reveals something you missed. \\
- Refine or strengthen your analysis based on this new information. Your updated analysis should be more nuanced. \\

You are arguing in favor of the image being authentic (real). \\
The other expert's (arguing for "AI-generated") points: \\

\{positive\_history\} \\

Provide your updated, refined analysis as a concise bullet-pointed list. \\
\end{tcolorbox}
\caption{Prompt template for the Con-Agent.}
\label{table:con_agent_prompt}
\end{table}

\subsection{Judge Agent Prompt}

The Judge Agent is tasked with overseeing the debate process and synthesizing the final decision based on both the tool-derived evidence and the debate history. 
The Judge also evaluates the sufficiency of each debate round and can decide to terminate the debate early if the arguments are deemed sufficient.
\autoref{table:judge_agent_prompt1} and \autoref{table:judge_agent_prompt2} are the prompt templates used to instruct the LLM:

\begin{table}[h!]
\centering
\begin{tcolorbox}[width=0.95\linewidth]
\small
As an impartial judge, review the debate history so far. \\
Your task is NOT to make the final decision, but to determine if the debate is sufficient to support a final decision. \\

Arguments for 'AI-generated': \\
\{positive\_args\} \\

Arguments for 'Authentic Image': \\
\{negative\_args\} \\

Your Decision Criteria: \\
1.  If one side's evidence is strong and the other's is weak or has been effectively countered, the information is likely sufficient. \\
2.  If both sides have presented compelling but conflicting evidence that has not yet been reconciled, more analysis is needed. \\
3.  If the discussion become repetitive, further rounds are unlikely to be productive. \\

Based on these criteria, decide if you have enough information to make a high-confidence final judgment. \\
Answer 'True' if sufficient, 'False' if more debate and analysis would be helpful. \\
\end{tcolorbox}
\caption{First prompt template for the Judge Agent.}
\label{table:judge_agent_prompt1}
\end{table}

\begin{table}[h!]
\centering
\begin{tcolorbox}[width=0.95\linewidth]
\small
You are an AI Image Forensics Judge. Your role is to synthesize all available information and deliver a definitive, well-reasoned judgment on whether the image is AI-generated or real. \\

- A real image refers to images created by humans, including photographs captured by cameras, photos that have been edited with software such as Photoshop, or human artistic creations such as hand-drawn sketches and paintings. \\
- An AI-generated image refers to images that are fully or partially generated by AI models. \\

Raw Evidence from tools:
{tool\_results}

Arguments for 'AI-generated': \\
\{positive\_args\} \\

Arguments for 'Authentic Image': \\
\{negative\_args\} \\

Your analysis must be a comprehensive synthesis. Follow these steps in your reasoning: \\
1.  Weigh the Evidence: Identify the most compelling piece of evidence from EACH side. \\
2.  Resolve the Core Conflict: Directly address the central disagreement. \\
3.  State Your Final Conclusion: Based on your analysis, provide a clear final verdict. \\

Required output format: \\
1. is\_ai\_generated: boolean (True if AI-generated, False if real image) \\
2. analysis\_details: A detailed analysis explaining your decision \\

Format the response as a structured object. \\
\end{tcolorbox}
\caption{Second prompt template for the Judge Agent.}
\label{table:judge_agent_prompt2}
\end{table}

\subsection{VLM Analysis Tool Prompt}

The VLM Analysis Tool utilizes vision-language models to conduct in-depth visual analysis of images. 
The prompt used to guide the VLM Analysis Tool is detailed in \autoref{table:vlm_analysis_tool_prompt}, ensuring the model focuses on key visual characteristics and provides comprehensive evidence to support its classification.

\begin{table}[h!]
\centering
\begin{tcolorbox}[width=0.95\linewidth]
\small
As a professional AI image detector, please analyze this image carefully: \\
        
1. Determine if this is an AI-generated image or a real image. \\
    - Real images include images that are created by humans, including photographs captured by cameras, photos that have been edited with software such as Photoshop, or human artistic creations such as hand-drawn sketches and paintings. \\
    - AI-generated images include images that are fully or partially generated by AI models. \\
            
2. If you determine it's an AI-generated image, please specifically identify and list the visual artifacts or characteristics that indicate AI generation, such as: \\
    - Unnatural textures or patterns \\
    - Inconsistent lighting or shadows \\
    - Anatomical errors in humans or animals \\
    - Unusual distortions or blending of elements \\
    - Text or writing abnormalities \\
    - Symmetry issues or repeating patterns \\
    - Unusual backgrounds or contextual inconsistencies \\
        
3. If you determine it's a real image, explain what characteristics support this conclusion. \\
        
4. Provide your final classification with confidence level (high, medium, or low). \\
\end{tcolorbox}
\caption{Prompt template for the VLM Analysis Tool.}
\label{table:vlm_analysis_tool_prompt}
\end{table}

\section{Metadata Analysis Tool Key Fields}
\label{appendix:metadata_fields}

\autoref{table:metadata_fields} provides the exact key fields and prefixes used in the metadata analysis tool to identify authenticity markers in images.

\begin{table*}[!t]
\centering
\caption{Metadata fields and prefixes considered in the analysis tool.}
\label{table:metadata_fields}
\setlength{\tabcolsep}{2pt}
\scalebox{0.80}{
\begin{tabularx}{\textwidth}{l|l|l}
\toprule
\textbf{Category} & \textbf{Field / Prefix} & \textbf{Description} \\
\midrule
\multicolumn{3}{c}{\textit{Exact Key Fields}} \\
\midrule
XMP:CreatorTool & Creator tool & Software used to generate or edit the image. \\
EXIF:Software & Software tag & Image editing or generation software information. \\
EXIF:UserComment & User comment & Arbitrary comments added to the image metadata. \\
File:Comment & File comment & Comments embedded directly in the file container. \\
XMP:Description & Description & Textual description of the image. \\
XMP:Title & Title & Title field embedded in XMP metadata. \\
XMP:Rights & Rights & Usage rights or copyright information. \\
XMP:Source & Source & Original source reference of the image. \\
EXIF:Make & Camera make & Manufacturer of the recording equipment. \\
EXIF:Model & Camera model & Camera model used for the photo. \\
EXIF:LensModel & Lens model & Lens information recorded by the camera. \\
EXIF:LensInfo & Lens info & Technical specifications of the lens. \\
EXIF:LensSerialNumber & Lens serial number & Unique identifier of the lens. \\
EXIF:ExposureTime & Exposure time & Shutter exposure duration. \\
EXIF:FNumber & F-number & Aperture size of the lens. \\
EXIF:ISO & ISO & Sensitivity setting of the camera. \\
EXIF:FocalLength & Focal length & Lens focal length value. \\
EXIF:SerialNumber & Camera serial number & Unique identifier of the camera. \\
EXIF:GPSLatitude & GPS latitude & Geographic latitude of capture. \\
EXIF:GPSLongitude & GPS longitude & Geographic longitude of capture. \\
EXIF:GPSTimeStamp & GPS timestamp & Time recorded by GPS. \\
EXIF:DateTimeOriginal & Original datetime & Original capture time of the image. \\
EXIF:CreateDate & Creation date & File creation date. \\
Composite:GPSPosition & GPS position & Combined GPS coordinates. \\
Composite:Aperture & Aperture & Derived aperture value. \\
Composite:ShutterSpeed & Shutter speed & Derived shutter speed. \\
Composite:LensID & Lens ID & Identifier for the lens model. \\
ICC\_Profile:ProfileDescription & ICC profile description & Description of the color profile. \\
ICC\_Profile:ProfileCopyright & ICC profile copyright & Copyright information for the ICC profile. \\
IPTC:DocumentNotes & Document notes & Notes in IPTC metadata. \\
IPTC:ApplicationRecordVersion & Record version & Version of the IPTC application record. \\
\midrule
\multicolumn{3}{c}{\textit{Key Field Prefixes}} \\
\midrule
MakerNotes:\* & Camera-specific notes & Manufacturer-specific EXIF metadata. \\
JUMBF:\* & JUMBF metadata & Metadata block for embedding auxiliary information. \\
MPF:\* & Multi-picture format & Metadata for multi-frame images. \\
\bottomrule
\end{tabularx}
}
\end{table*}

\section{List of Selected Models from Hugging Face}
\label{appendix:hf_models}

We select the top five most downloaded classification models for AI-generated image detection available on Hugging Face:

\begin{itemize}
    \item haywoodsloan/ai-image-detector-deploy
    \item Organika/sdxl-detector
    \item legekka/AI-Anime-Image-Detector-ViT
    \item Smogy/SMOGY-Ai-images-detector
    \item NYUAD-ComNets/NYUAD\_AI-generated\_images\_detector
\end{itemize}

All models are publicly available on the Hugging Face Hub.

\section{AI Model Sources}
\label{appendix:ai_model_sources}

\autoref{table:ai_model_sources} provides an overview of the AI models used for generating images in our benchmark's AI-sourced datasets.

\begin{table*}[!t]
\centering
\caption{Overview of the AI models and platforms used for generating images in our benchmark's AI-sourced datasets. 
The table is categorized by the \textit{In-the-Lab} and \textit{In-the-Wild} settings.}
\label{table:ai_model_sources}
\scalebox{0.80}{ 
\begin{tabularx}{\linewidth}{l X}
\toprule
\textbf{Dataset Source} & \textbf{AI Models and Platforms Used for Generation} \\
\midrule
\multicolumn{2}{c}{\textbf{In-the-Lab AI Image Sources}} \\
\midrule
GenImage~\cite{ZCYHLLTHHW23} & BigGAN, GLIDE, VQDM, ADM, Midjourney, Wukong, and Stable Diffusion (v1.4, v1.5). \\
\addlinespace 
FakeBench~\cite{LLWLWRL24} & ProGAN, StyleGANs, CogView2, FuseDream, VQDM, GLIDE, Midjourney, Stable Diffusion, DALL·E 2, and DALL·E 3. \\
\midrule
\multicolumn{2}{c}{\textbf{In-the-Wild AI Image Sources}} \\
\midrule
Lexica & Lexica Aperture Series (v3.5, v4, v5, Max). \\
\addlinespace
Nightcafe & DALL·E 2, DALL·E 3, Stable Diffusion, and various other community fine-tuned models. \\
\addlinespace
Civitai & A vast collection of community fine-tuned models, predominantly based on Stable Diffusion (including SDXL variants) series. \\
\bottomrule
\end{tabularx}
}
\end{table*}

\section{Detailed Accuracy Performance}
\label{appendix:detailed_accuracy}

\autoref{table:detailed_accuracy} provides a detailed breakdown of the accuracy performance of various methods across different image sources, including both in-the-lab and in-the-wild scenarios.

\begin{table*}[!t]
\centering
\caption{Detailed Accuracy performance of different methods on each image source.}
\label{table:detailed_accuracy}
\setlength{\tabcolsep}{3pt} 
\scalebox{0.80}{ 
\begin{tabular}{l ccc cc @{\hskip 12pt} ccc ccc}
\toprule
& \multicolumn{5}{c}{\textbf{In-the-Lab}} & \multicolumn{6}{c}{\textbf{In-the-Wild}} \\
\cmidrule(lr){2-6} \cmidrule(lr){7-12}
& \multicolumn{3}{c}{Real Images} & \multicolumn{2}{c}{AI Images} & \multicolumn{3}{c}{Real Images} & \multicolumn{3}{c}{AI Images} \\
\cmidrule(lr){2-4} \cmidrule(lr){5-6} \cmidrule(lr){7-9} \cmidrule(lr){10-12}
\textbf{Method} & Flickr30k & ImageNet & DIV2K & GenImage & FakeBench & Holopix50k & Flickr & W. Commons & Lexica & Nightcafe & Civitai \\
\midrule
CNNSpot~\cite{WWZOE20} & 0.9980 & 0.9980 & 1.0000 & 0.0475 & 0.1857 & 1.0000 & 1.0000 & 0.9980 & 0.0000 & 0.0020 & 0.0000 \\
PatchCraft~\cite{ZXLQZ24} & 0.9900 & 0.9140 & 0.7680 & 0.8525 & 0.5986 & 0.8600 & 0.9040 & 0.5960 & 0.0140 & 0.2360 & 0.3360 \\
DE-FAKE~\cite{SLYZ22} & 0.8980 & 0.7280 & 0.3480 & 0.7225 & 0.6443 & 0.7180 & 0.5700 & 0.4920 & 0.9840 & 0.9380 & 0.8360 \\
GPT-4o~\cite{GPT4o} & 1.0000 & 0.9860 & 0.9960 & 0.8850 & 0.9471 & 0.9980 & 0.9820 & 0.9940 & 0.7520 & 0.9900 & 0.9420 \\
AIFo (ours) & 1.0000 & 0.9840 & 0.9920 & 0.9475 & 0.9657 & 0.9960 & 0.9880 & 0.9940 & 0.8420 & 0.9880 & 0.9940 \\
\bottomrule
\end{tabular}
}
\end{table*}

\end{document}